\documentclass[runningheads]{llncs}

\usepackage{eccv}

\usepackage{eccvabbrv}

\usepackage{graphicx}
\usepackage{booktabs}
\usepackage{algorithm}
\usepackage{algorithmicx}
\usepackage{algpseudocode}
\usepackage{amssymb}
\usepackage{amsmath}
\usepackage{graphicx}
\usepackage{multirow}
\usepackage{wrapfig}
\usepackage{float}

\usepackage[accsupp]{axessibility}  

\usepackage{hyperref}

\usepackage{orcidlink}

\begin{document}

\title{Text-Anchored Score Composition: Tackling Condition Misalignment in Text-to-Image Diffusion Models} 

\titlerunning{Text-Anchored Score Composition}

\author{
Luozhou Wang\inst{1}\thanks{Equal contribution} \and
Guibao Shen\inst{1}\protect\footnotemark[1] \and
Wenhang Ge\inst{1} \and
Guangyong Chen\inst{3,4} \and \\
Yijun Li\inst{5} \and
Yingcong Chen\inst{1,2}\thanks{Corresponding author}
}
\authorrunning{L. Wang et al.}

\institute{
Hong Kong University of Science and Technology (Guangzhou) \and 
Hong Kong University of Science and Technology \and 
ZhejiangLab \and 
Zhejiang University \and 
Adobe Research
}

\maketitle
\begin{figure}[ht]
\centering
\includegraphics[width = \linewidth]{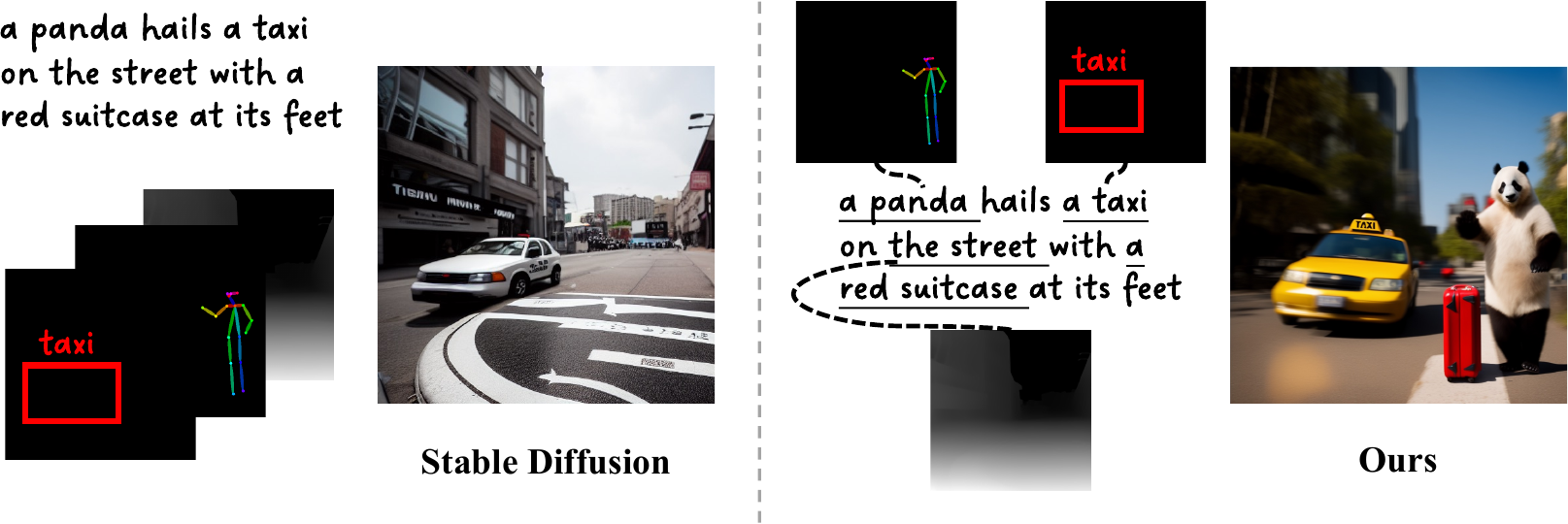}
        \captionof{figure}{Illustration of our proposed \textbf{Text-Anchored Score Composition} showcasing the ability to handle the misalignment between conditions for controllable generation tasks. For example, with several control conditions (e.g., depth, pose, bounding box) specifying the layout structure and the text condition indicating \textbf{extra} guidance (e.g., new object, new correspondence of spatial relationship), our method is able to generate high-quality plausible outputs that satisfy all given conditions without training.}
        \label{fig-teaser}
\end{figure}

\begin{abstract}
Text-to-image diffusion models have advanced towards more controllable generation via supporting various additional conditions (e.g., depth map, bounding box) beyond text.
 However, these models are learned based on the premise of perfect alignment between the text and extra conditions. If this alignment is not satisfied, the final output could be either dominated by one condition, or ambiguity may arise, failing to meet user expectations.
To address this issue, we present a training-free approach called \textbf{Text-Anchored Score Composition (TASC)} to further improve the controllability of existing models when provided with partially aligned conditions. 
The TASC firstly separates conditions based on pair relationships, computing the result individually for each pair.
This ensures that each pair no longer has conflicting conditions. 
Then we propose an attention realignment operation to realign these independently calculated results via a cross-attention mechanism to avoid new conflicts when combining them back. 
Both qualitative and quantitative results demonstrate the effectiveness of our approach in handling unaligned conditions, which performs favorably against recent methods and more importantly adds flexibility to the controllable image generation process.
\end{abstract}

\section{Introduction}
\label{sec:intro}

Diffusion models ~\cite{sohl2015deep,ho2020denoising,song2020ddim,dhariwal2021guided,nichol2021glide,saharia2022imagen,ramesh2022dalle2,rombach2022high, nichol2021glide, saharia2022imagen, ramesh2022dalle2}, epitomized by Stable Diffusion (SD)~\cite{rombach2022high}, are notably proficient in image synthesis. Central to their functioning, and a principle they share with score-based models \cite{song2019score1,song2020score2}, is the prediction of the underlying score of the data distribution. Building upon this, advancements have been made by extending SD with additional conditions, specifically through the implementation of ControlNet~\cite{zhang2023adding}, the T2I Adapter~\cite{mou2023t2i} and GLIGEN \cite{li2023gligen}. These innovations enhance controllability with the idea of the adapter, thereby obviating the need for training from scratch.

\begin{figure}[t]
    \centering
    \includegraphics[width=1.0\linewidth]{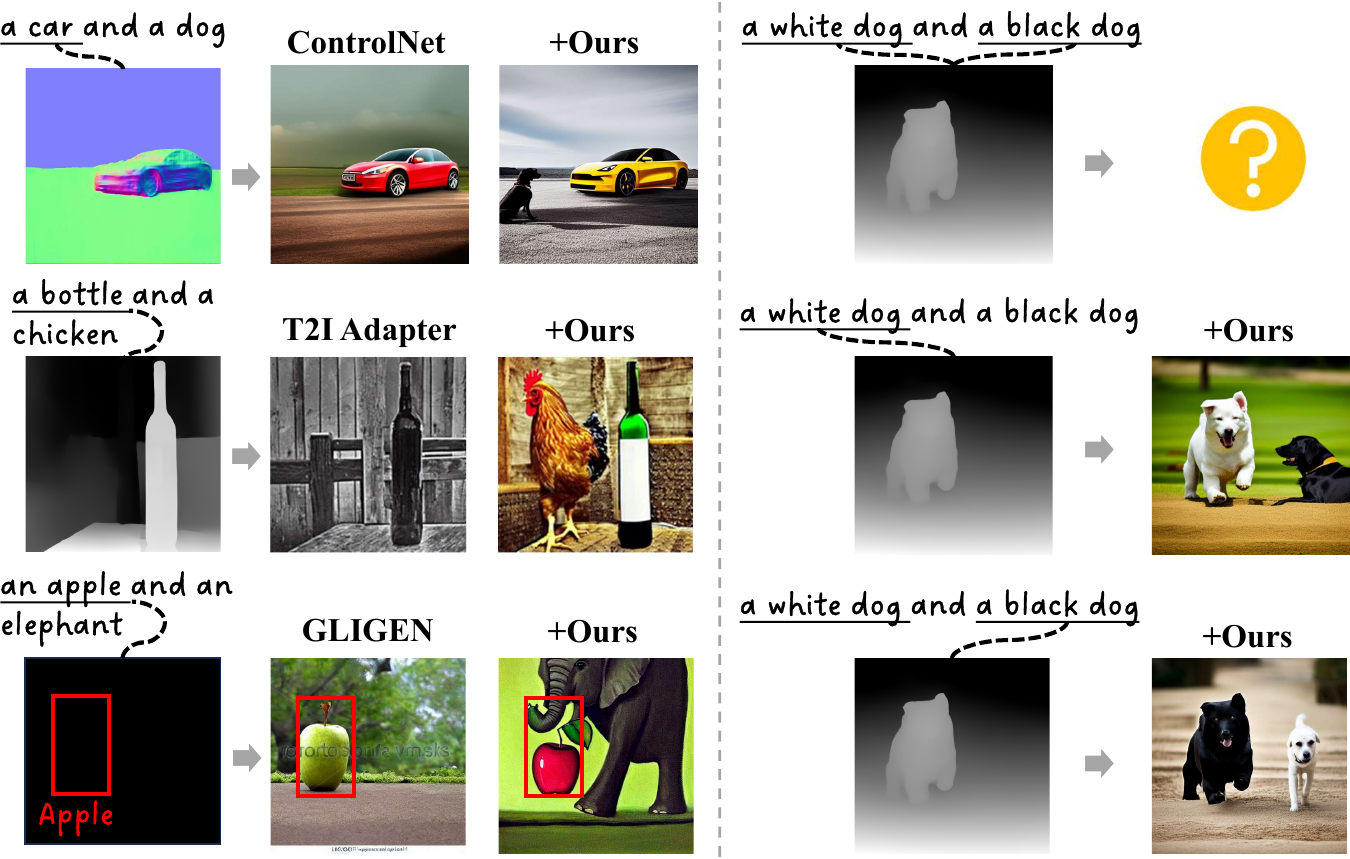}
    \caption{\textbf{Challenges in Multi-Condition Image Synthesis.} Left: the ``dominance'' effect exists across various methods, where one object (\eg, car) overshadows the generation, omitting the other (\eg, dog). Right: The ``ambiguity'' issue, where it's unclear which object from the text matches the intended depth condition.}
    \label{fig-intro}
\end{figure}

However, the increased complexity of control signals originates another challenge: multi-condition controllable image generation relies on the assumption that all conditions are well-aligned. This strict assumption limits the scenarios in which users can effectively employ controllable generation.  
The misalignment of multiple control conditions can negatively impact the overall performance and user experience during controllable image generation. This misalignment typically results in two phenomena: (i) \emph{Dominance}, where one condition takes over the output; (ii) \emph{Ambiguity}, leading to unclear or conflicting outcomes.
For instance, consider the case where the text condition mentions two objects, but the extra condition contains only one object. Dominance occurs when the extra condition accurately corresponds to one of the objects in the text, leading the generation process to be dominated by the extra condition and neglecting the other object. As illustrated on the left side of Figure~\ref{fig-intro}, the dog in the text ``a car and a dog'' is omitted. This phenomenon happens across different control methods \cite{zhang2023adding, mou2023t2i, li2023gligen}.
On the other hand, ambiguity arises when the extra condition could refer to either one of the objects in the text, resulting in an unclear correspondence. In this situation, it becomes challenging to successfully generate an image where the extra condition accurately corresponds to a specific object mentioned in the text, as designated by the user. As depicted on the right side of Figure~\ref{fig-intro}, The user is unable to control which object to generate simply based on the depth map.



In this work, we introduce a training-free approach to address the misalignment between text and image conditions for a broader range of controllable image generation needs. 
We propose the \textbf{Text-Anchored Score Composition (TASC)}, which treats the entire text as a unified condition, with each additional condition corresponding to a different subset of words in the text.
Firstly, our score composition formula separates conditions based on pair relationships by leveraging the fine-grained needs of users. 
During each timestep of the diffusion generation process, rather than inputting all conditions together into the denoising network, we calculate the model output for each condition pair separately, termed the ``individual score.'' 
Conversely, the ``unified score'' is computed using the entire text input as a single condition. 
This process ensures that the impact of each condition remains unaffected by the dominance of other conditions. 
Nevertheless, the computation process of these scores can be further optimized. 
We observed that at each step of diffusion, the calculations for individual scores and the unified score are independent, potentially leading to conflicts between the unified and individual scores. 
To address this, we propose an attention alignment operation, aimed at resolving the conflict between the unified and individual scores. 
This operation realigns the attention values of the unified score with those of the individual scores, resulting in a modified unified score. 
This modified unified score is then combined with the individual scores and used for diffusion sampling to produce the final output.

Our Text-Anchored Score Composition (TASC) offers a versatile and comprehensive solution for controllable image generation, addressing various needs while reducing potential conflicts and overlaps between conditions as shown in Figure~\ref{fig-teaser}. In conclusion, our contributions can be summarized as follows:
\begin{itemize}
    \item We aim at enabling more flexible controllable generation when provided with partially aligned conditions, which is significantly different from the conventional need for perfectly aligned conditions in previous works.

    \item We propose a training-free approach and demonstrate its effectiveness with extensive experiments to efficiently compute and align both individual and unified score estimates, 
    generating high-quality results that meet user expectations.
\end{itemize}

\section{Related Work}
\label{sec:related_work}
\noindent\textbf{Text-to-Image Diffusion.}~Diffusion model has emerged as a powerful technique for generating high-quality images~\cite{rombach2022high,yang2023denoising,nichol2021glide,ramesh2022dalle2,saharia2022imagen}. The Stable Diffusion (SD)~\cite{rombach2022high} model, which leverages latent diffusion processes, serves as a prime example. 
This text-to-image diffusion model becomes the foundation for generation of other modality and tasks~\cite{liang2023luciddreamer,poole2022dreamfusion,wang2023modelscope,wang2023not}.
Further, certain studies~\cite{ruiz2022dreambooth, gal2022image, hu2021lora, huang2023reversion} have expanded on this foundation, incorporating image guidance into the diffusion process for customization. By binding specific image features with the text, these models simulate the corresponding style through the text prompt. Others~\cite{choi2021ilvr,brack2023sega,meng2021sdedit,zhao2022egsde, zeng2022scenecomposer}  focus on exploiting spatial information inherent in images, integrating the image directly into the computations to ensure better alignment in the final output.

\noindent\textbf{Multi-Condition Control.}~The ability to integrate multiple control conditions into existing models has been a notable advancement, facilitated by techniques like ControlNet~\cite{zhang2023adding} or T2I adapter~\cite{mou2023t2i}. These methods enable diffusion models like Stable Diffusion to handle multiple image conditions without the need for extensive retraining. 
These methods support a variety of condition types, such as depth, normal, human pose, and canny, thereby expanding the scope of image generation tasks. However, the introduction of multiple conditions leads to the issue of condition composition. \textbf{Although previous works, like~\cite{liu2022compositional,brooks2022instructpix2pix,bhunia2022person,wang2024instancediffusion}, have addressed the composition of multiple conditions during generations, none of them have discussed the potential misalignment among these conditions.} Building upon these approaches, our work is the first attempt to address this problem, providing a training-free approach to unify the misaligned conditions, and thus significantly improve their controllability.

\noindent\textbf{Cross-Attention Modifications.}~The cross-attention mechanism has become a cornerstone of the text-to-image diffusion model, with several studies~\cite{hertz2022prompt, mokady2022null, tumanyan2022plug, parmar2023zero} employing this mechanism to execute a variety of image editing tasks. Notably, altering attention values has been suggested as an effective strategy for steering image generation, as exemplified in models such as StructureDiffusion~\cite{feng2022training} and Attend-and-Excite ~\cite{chefer2023attend}. 
During the generation process, Attend-and-Excite ~\cite{chefer2023attend} modifies the latent at each timestep by maximizing the attention value of designated tokens, thereby ensuring the synthesis of both objects in the final output. Conversely, StructureDiffusion~\cite{feng2022training} disassembles the text input into hierarchical levels and consolidates them through cross-attention computation. 
Several studies have also employed cross-attention mechanisms to precisely regulate the positioning of specific contents or objects within images~\cite{chen2024training, xie2023boxdiff, ge2023expressive, kim2023dense}.
Nevertheless, these methodologies exhibit diminished efficacy in addressing the dominance issue when confronted with multiple conditions. More critically, they often stumble upon ambiguity, as illustrated in Figure~\ref{fig-intro}, where users are incapable of controlling the specific region for object generation. 

\section{Proposed Method}
\label{sec:methods}

\subsection{Preliminary}


\noindent\textbf{Multiple-Conditioned Image Generation.}~
Our multi-condition framework requires additional parameters, such as distinct sets $\phi$ for different conditions (e.g., ControlNet \cite{zhang2023adding} or GLIGEN \cite{li2023gligen}). When computing scores, it's crucial to specify the active conditions, even if some are inactive. To streamline notation for active conditions, we use simplified expressions. For instance, $\epsilon_{\theta,\phi_2}(z_t, \varnothing, \mathcal{I})$ represents a scenario with only the additional condition $\mathcal{I}$ active, maintaining the text condition slot in our notation regardless of use. This approach helps distinguish our method from others like Composable Diffusion \cite{liu2022compositional}, which doesn't fully consider the complex interactions between conditions and text in multi-condition settings.


\subsection{Condition Misalignment}
\begin{figure*}[t]
    \centering
    \includegraphics[width=1.0\linewidth]{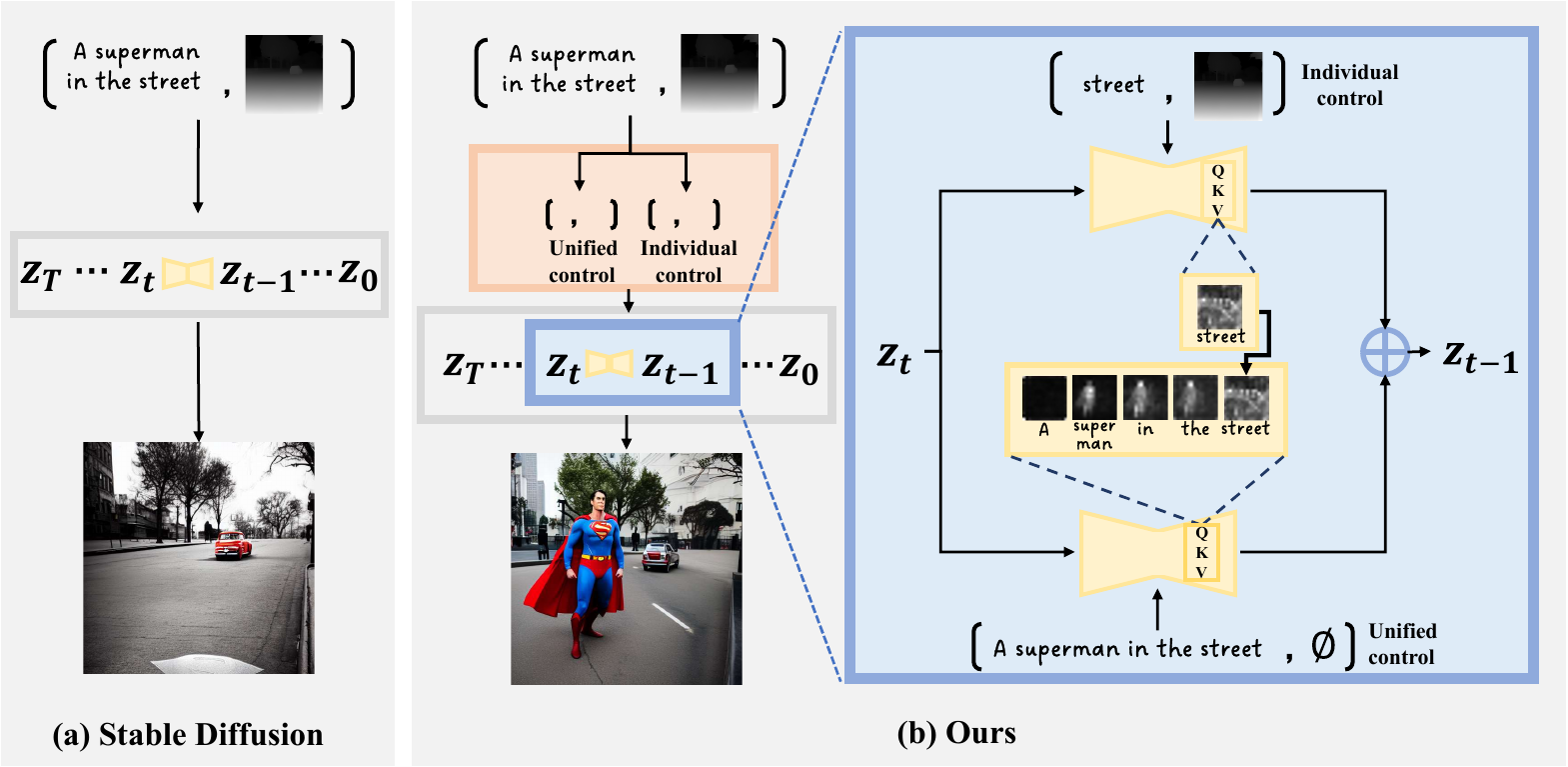}
    \caption{\textbf{Image Synthesis using Text-Anchored Score Composition.} Given a depth map of a street and a textual description specifying ``a Superman in the street'', our method effectively generates the corresponding image.}
    \label{fig-method}
\end{figure*}

Given a text prompt $\mathcal{P}$ and additional conditions $\mathbb{I}=[\mathcal{I}_1, \mathcal{I}_2,\dots, \mathcal{I}_K]$, the score of the original classifier-free guidance~\cite{ho2022classifier} is computed as follows:

\begin{align}
& \epsilon_{\text{CFG}}(z_t, \mathcal{P}, \mathbb{I}) = \epsilon_\theta(z_t, \varnothing, \varnothing) \nonumber \\ 
& + w \cdot (\epsilon_{\theta,\phi_{1,\dots,K}}(z_t, \mathcal{P}, \mathbb{I}) - \epsilon_\theta(z_t, \varnothing, \varnothing)).
\label{eq-base-cfg}
\end{align}

Often times we feed conditions into the model collectively, without addressing potential misalignment issues, \ie, the dominance and ambiguity. 
We introduce and analyze these two phenomena through the example with $K=1$ as shown in Figure~\ref{fig-intro}.
\textbf{Dominance} represents a form of partial alignment, where additional condition $\mathcal{I}_1$ corresponds solely to one object $\mathcal{P}_i$ in the overall text prompt $\mathcal{P}$. However, when these conditions are simultaneously fed into the model, it struggles to concurrently fulfill the text and additional conditions. Consequently, the output exhibits a dominance phenomenon, with one object receiving precedence while the others are disregarded.
\textbf{Ambiguity} emerges as a by-product of dominance, wherein the additional condition might correspond to any object within the text, \ie, $\mathcal{I}_1$ could either correspond to $\mathcal{P}_i$ or $\mathcal{P}_j$. Thus the conditions fed into the model create a potential for ambiguity. Moreover, in practical applications, even when users delineate the correspondence, current methods find it challenging to leverage this information effectively.

\subsection{Our Solution}
To address the problem discussed earlier, we propose a Text-Anchored Score Composition. Our solution consists of two components: the \textit{Text-Anchored Score Composition formula} , which breaks down the conditions into fully-aligned pairs, and the \textit{attention realignment}, which ensures that the decomposed results are more consistent when combined together.

\noindent\textbf{Text-Anchored Score Composition.}
Initially, we assume users will specify pair relationships \( S(\cdot) \) to setting the details of individual entities in the text, using image conditions or bounding boxes, like setting a dog's color as shown in Figure~\ref{fig-intro}. 
We presume users won't provide meaningless inputs, ensuring pair relationships are practical and aligned with their creative goals.

This function formalizes the pairing between additional conditions and text tokens.
By this function, we can establish aligned condition pairs $\{(\mathcal{P}_{S(k)}, \mathcal{I}_k)\}_{k=1}^K$.
Leveraging these pairs, we introduce the \textbf{Text-Anchored Score Composition} (TASC) equation:

\begin{align}
& \epsilon(z_t, \mathcal{P}, \mathbb{I}) = \epsilon_\theta(z_t, \varnothing, \varnothing) \nonumber \\ 
& + w_0 \cdot \underbrace{(\epsilon_\theta(z_t, \mathcal{P}, \varnothing) - \epsilon_\theta(z_t, \varnothing, \varnothing))}_{\text{unified control}} \nonumber \\
& + \sum_{k=1}^K w_k \underbrace{(\epsilon_{\theta,\phi_k}(z_t, \mathcal{P}_{S(k)}, \mathcal{I}_k) - \epsilon_\theta(z_t, \varnothing, \varnothing))}_{\text{individual control}}.
\label{eq-dr-cfg}
\end{align}

We ensure the preservation of the original text $\mathcal{P}$ as the sole input, serving as a unified control. 
This strategy ensures the successful generation of all objects, even those without corresponding additional conditions. 
Note if without these pair relationships \( S(\cdot) \), our method just defaults to standard image-conditioned generation.

Through the equation, we generate individual scores for each pair, eliminating any existing misalignment issues within the computation of individual scores. 
Both text and additional conditions, therefore, contribute to the final image. 
Additionally, the implementation of the pair relationship function $S(\cdot)$ allows us to further mitigate any ambiguity based on the specific needs of the user.

\begin{figure}[t]
    \centering
    \includegraphics[width=1.0\linewidth]{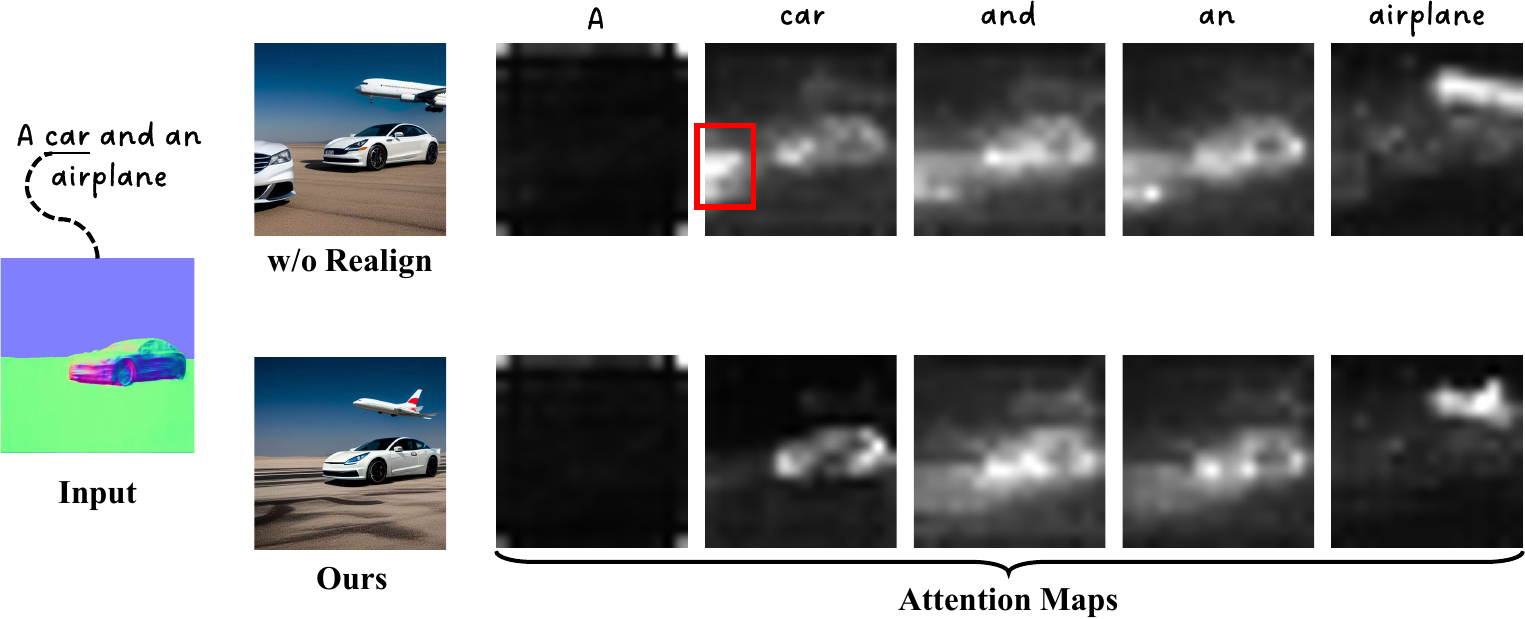}
    \caption{\textbf{Illustration of the ``Attention Realignment'' operation.} The operation addresses mismatches between the unified score and individual scores during generation, preventing the final object from appearing multiple times,  as individual score for the same token ``car'' will not contribute to the attention value in the red box.}
    \label{fig-method-motivation}
\end{figure}

However, this equation introduces a potential problem: the unified score and individual scores are calculated independently, which may lead to discrepancies between them. 
Consequently, the synthesized image might contain some undesirable content, as depicted in Figure~\ref{fig-method-motivation}.
By analyzing the attention map during the generation process, we find that the attention map for the ``car'', which is already constrained by the depth map condition, also has high values in other areas. 
Intuitively, when generating using individual scores and unified scores separately, the value corresponding to the ``car'' token on the attention map in the individual score always aligns with the depth map condition, thus restricting it to the correct area. 
Conversely, the ``car'' in the unified score is unconstrained. 
We therefore deduce that this phenomenon is due to the lack of consensus between the unified score and the individual scores on the same token.

\noindent\textbf{Attention Realignment.}~To ensure appropriate alignment between the unified score and the corresponding individual scores, we introduce an attention realignment operation. During the computation of the individual and unified scores, we can obtain the corresponding attention maps $M^0$ and $\{M^k\}_{k=1}^K$.
We then replace the attention values of the corresponding tokens \( S(k) \) with the attention values derived from the \( k \)-th individual score. Specifically, \(M^0_{S(k)} \leftarrow M^k\) for \(k=1,\dots,K\), where \(M^0_{S(k)}\) is the attention map of the \(S(k)\)-th token in \(\epsilon_\theta(z_t, \mathcal{P}, \varnothing)\), and \(M^k\) is that of \(\epsilon_{\theta,\phi_k}(z_t, \mathcal{P}_{S(k)}, \mathcal{I}_k)\). This leads to aligned unified control.

We then substitute this $\epsilon_\theta(z_t, \mathcal{P}, \varnothing)$ into the unified control of Eq.~\eqref{eq-dr-cfg} to obtain our final score $\tilde{\epsilon}$, which is subsequently used for sampling. 
The modified unified score can prevent the generation of undesirable content during the generation process. Additionally, the motivation behind replacing attention is to ensure appropriate interactions; that is, once the car's information is confirmed, the generation of the airplane must be reasonable and should not occupy the car's attention. 

Detailed descriptions and mathematical formulations of these operations are provided in the supplementary material.
The entire procedure is defined in Algorithm~\ref{alg:cap}.
\begin{algorithm}[!htbp]
\caption{Text-Anchored Score Composition.}\label{alg:cap}
\begin{algorithmic}[1]
\State {\bfseries Input:} unified text prompt $\mathcal{P}$, additional conditions $\mathbb{I}$, pair relationship function $S(\cdot)$, pretrained diffusion model with $K$ controller model parameters $\epsilon_{\theta, \phi_{1,\dots,K}}$
\State Initialize sample $z_T \sim N (0, 1)$
\For{$t=T$ to $1$}
    \State $\epsilon, \_ \leftarrow \epsilon_{\theta}(z_t, \varnothing, \varnothing)$
    \For{$k=1$ to $K$}
        \State $\epsilon^k, M^k \leftarrow \epsilon_{\theta,\phi_k}(z_t, \mathcal{P}_{S(k)}, \mathcal{I}_k)$
    \EndFor
    \State $\epsilon^0, \_ \leftarrow \epsilon_{\theta}(z_t, \mathcal{P}, \varnothing)\{M^k\}^K_{k=1}$ \Comment{attention realignment using $\{M^k\}^K_{k=1}$}
    \State $\tilde{\epsilon} = \epsilon + w_0 \cdot (\epsilon^0 - \epsilon) + \sum_{k=1}^K w_k \cdot (\epsilon^k - \epsilon) $
    \State $z_{t-1} \leftarrow \tilde{\epsilon}, z_t$ \Comment{sampling}
\EndFor
\end{algorithmic}
\end{algorithm}

\noindent\textbf{Discussions.}
Our method effectively processes complex text prompts and a variety of image control signals, as illustrated in Fig.~\ref{fig-teaser}. 
When calculating a unified score for lengthy text prompts, the issue of missing objects may still arise. 
Nonetheless, our approach is fully compatible with existing solutions for missing objects, such as Attend \& Excite~\cite{chefer2023attend}, demonstrating its adaptability and ensuring comprehensive object generation.

\begin{figure*}[t]
    \centering
    \includegraphics[width=1.0\linewidth]{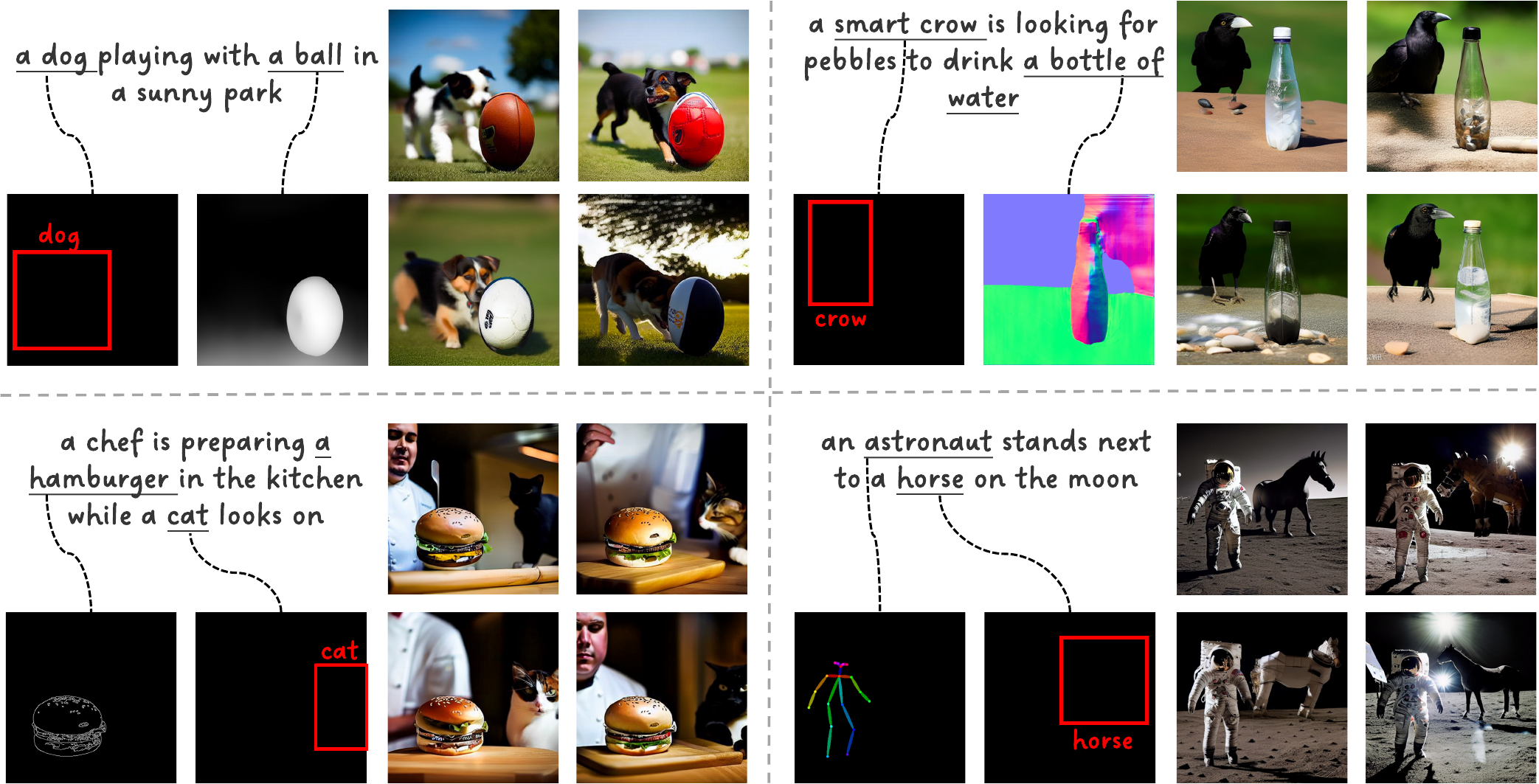}
    \caption{\textbf{Applications.} Demonstrating scalability, our method adeptly manages complex scenes, seamlessly combining controller mechanisms like both ControlNet \cite{zhang2023adding} and GLIGEN \cite{li2023gligen} simultaneously.} 
    \label{fig-exp-complex_scenes}
\end{figure*}

Furthermore, we recognize that condition conflicts can arise with various adapter combinations, including but not limited to ControlNet~\cite{zhang2023adding}, GLIGEN~\cite{li2023gligen}, and T2I Adapters~\cite{mou2023t2i}. 
Our novel score composition formula, presented in Equation~\eqref{eq-dr-cfg}, allows us to effectively address these conflicts, as demonstrated in Fig.~\ref{fig-exp-complex_scenes}.

\section{Experimental Results}
\label{sec:experiments}

\begin{figure*}[t]
    \centering
    \includegraphics[width=1.0\linewidth]{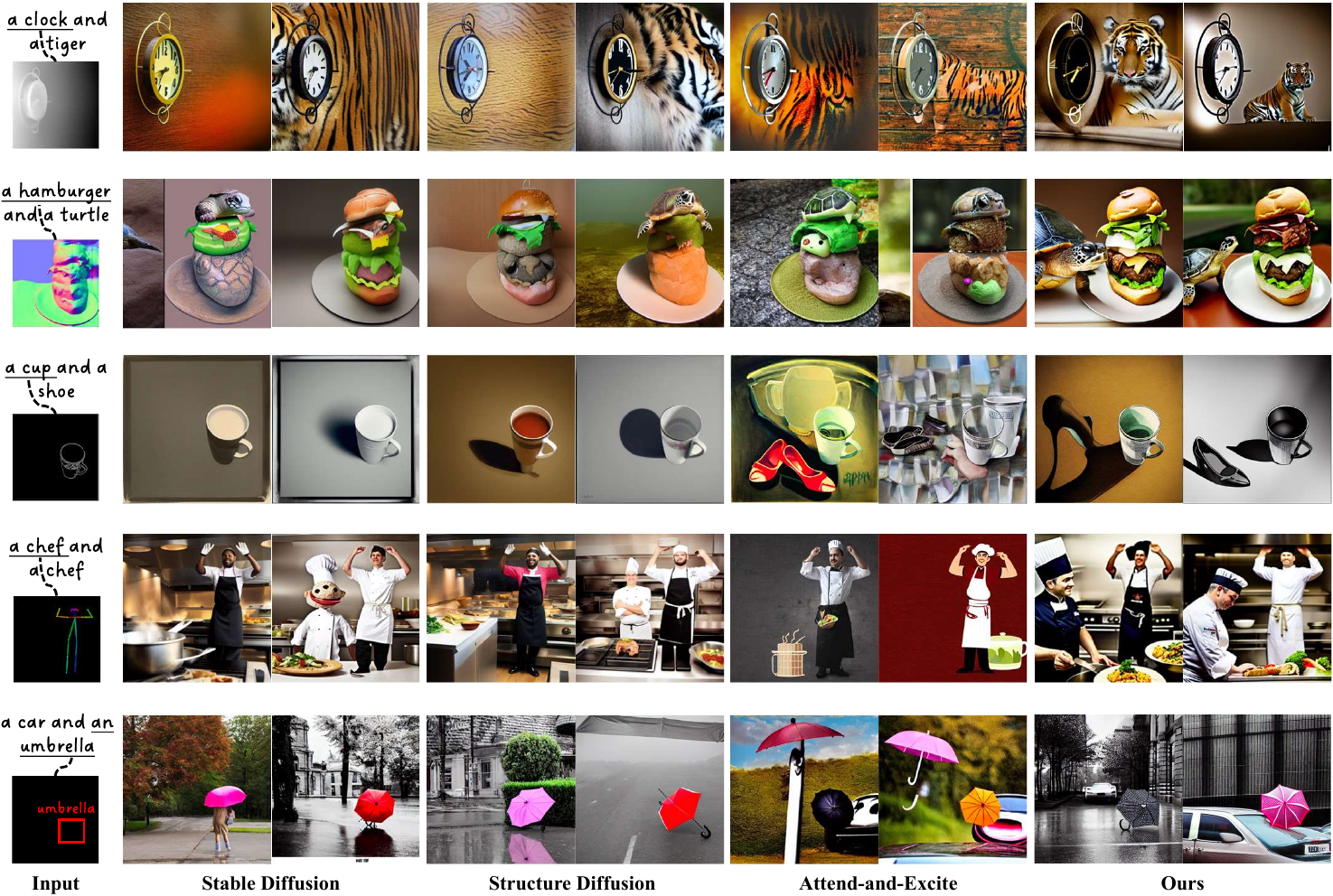}
    \caption{\textbf{Comparisons of ``Dominance'' Challenge.} Our approach ensures that objects in the text, not linked to additional conditions, are still accurately generated (\eg, ``tiger'' in the first example).}
    \label{fig-exp-dominance}
\end{figure*}

\begin{figure*}[t]
    \centering
    \includegraphics[width=1.0\linewidth]{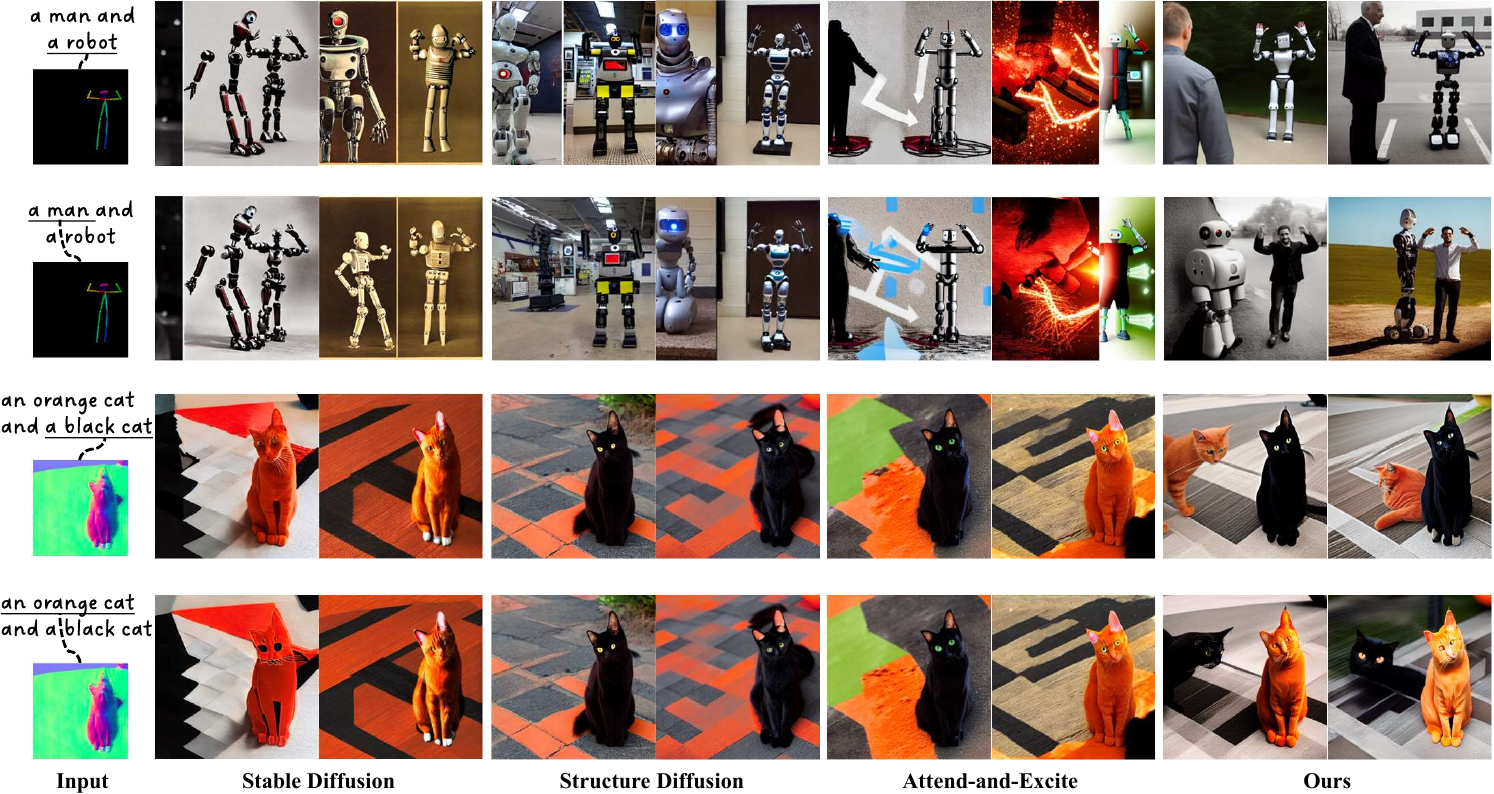}
    \caption{\textbf{Comparisons of ``Ambiguity'' Challenge.} Our method accurately maps specific words in the text (\eg, ``orange cat'' in the last row) to the corresponding additional conditions (``normal map'').}
    \label{fig-exp-ambiguity}
\end{figure*}

\subsection{Evaluation Setup}
Our experiments employed widely-used controllable methods, specifically those in \cite{zhang2023adding, mou2023t2i, li2023gligen}. Given the similar functionalities of the T2I adapter \cite{mou2023t2i} and ControlNets \cite{zhang2023adding}, we chose ControlNets for our study. The image conditions primarily included depth, normal, canny, and pose. Additionally, we incorporated grounding token-controlled image generation as described in \cite{li2023gligen}. The evaluation set, following \cite{chefer2023attend}, comprised prompts like ``a [\emph{objectA}] and a [\emph{objectB}].'' Each prompt included a control condition for one object, with the other generated textually. 
Moreover, testing scenarios include specialized prompts for ``pose'' evaluations, such as \textbf{``a [character] and an [object]'' and ``a [character A] and a [character B],''} with each character depicted in a variety of poses. 
In total, 444 and 448 unique prompts were used for image conditions and grounding tokens, generating 10 images each.
Consider the diversity of prompts, we have significantly expanded the \textbf{diversity to include various control types and control images}, in addition to textual prompts.

\subsection{Qualitative Evaluation}

\noindent\textbf{Applications.}~As seen in Figure~\ref{fig-exp-complex_scenes}, our method proves effective in complex scenes with various conditions. Take the intricate prompt ``a smart crow is looking for pebbles to drink a bottle of water,'' involving three entities (crow, pebbles, bottle) and control signals for two (crow, bottle). Our method adeptly generates all three entities. Importantly, in multi-condition scenarios, conflicts can emerge both between extra conditions (like control image and grounding token) and the text, as well as among the conditions themselves, heightening the complexity of these scenes.

\noindent\textbf{Comparisons of ``dominance'' Challenge.}~The results depicted in Figure~\ref{fig-exp-dominance} illustrate our method's effectiveness. It reliably generates every object described in the text across diverse controllable methods, surpassing baseline techniques. Moreover, it maintains a coherent and harmonious composition within each scene.

\noindent\textbf{Comparisons of ``ambiguity'' Challenge.}~Our method effectively binds condition to one of the text elements, thereby eliminating this phenomenon. While for baseline methods, which lack this capability, we rewrite the text to reduce ambiguity. For example, given an image with a pose on the right, we add directional words to the text, like ``a man on the left and a robot on the right.'' However, even with text modification, other methods struggle to accurately match the specifics, whereas our method correctly assigns the specified poses to the characters in the text, as shown in Figure~\ref{fig-exp-ambiguity}.

\subsection{Quantitative Evaluation}

\begin{table}[!ht]
    \caption{\textbf{Quantitative Comparison Results.}}
    \centering
    \resizebox{1.0\columnwidth}{!}{%
    \begin{tabular}{cccccc}
    \toprule
        \textbf{Method} & \textbf{CLIP Score} & \textbf{BLIP Score}  &  \textbf{FID}&\textbf{Time
(Seconds)}&\textbf{Memory
(MB)}\\ \midrule
        SD & 0.2934 & 0.6949  &  112.89&\textbf{9.70}&\textbf{7260.0}\\ 
        SSD & 0.2928 & 0.6940  &  112.57&10.87&8730.0\\ 
        A\&E & 0.3253 & 0.7334  &  111.41&19.36&18840.2\\ \midrule
        \textbf{Ours}  & \textbf{0.3323} & \textbf{0.7388}  &  \textbf{107.39}&13.88&9764.0\\ \bottomrule
    \end{tabular}
    }
    \label{tab:adapter}
\end{table}

\begin{table}[tb]
\caption{\textbf{Impact on Controllability with Different Methods.} Low is better.}
\centering
\resizebox{0.7\columnwidth}{!}{%
\begin{tabular}{ccc}
\toprule
                        & with ControlNets\cite{zhang2023adding}    & with GLIGEN\cite{li2023gligen}  
                                                                            \\ \midrule                    
{A\&E \cite{chefer2023attend}}     
                        & 14.1\%                                    & 7.36\% \\
{Ours}
                        & \textbf{ 1.6\%}                           & \textbf{4.34\%} \\ \bottomrule
\end{tabular}%
}
\label{tab-image-condition}
\end{table}

\noindent\textbf{Image-text similarity.}~ For textual alignment, we compute the cosine similarity between the input text prompt and the generated image, which is called the CLIP Score. Furthermore, We pivot to generate matching image captions using a pre-trained BLIP~\cite{li2022blip} and then compute the average CLIP similarity between the original prompt and all generated captions, which is called BLIP score. Both of the metrics are the higher the better.
Table~\ref{tab:adapter} presents our quantitative results. We demonstrate superior performance across different controllable methods. For StructureDiffusion~\cite{feng2022training}, we observe a marginal decrease in metrics compared to Stable Diffusion, an observation congruent with the findings in ~\cite{chefer2023attend}.

\noindent\textbf{Image Quality and Diversity.}~ For image Quality and diversity, we adopt the widely used Fréchet Inception Distance(FID)\cite{heusel2017gans} tested on MSCOCO dataset\cite{lin2014microsoft} for evaluation. It computes the distance between the distribution of the synthetic images and that of the real test images dataset. The FID is better when lower. Table~\ref{tab:adapter} demonstrates that our proposed method outperforms other baseline methods by a significant margin in terms of image quality.

\noindent\textbf{Computation Cost.}~ We also report both GPU memory usage and inference time of each method in Table~\ref{tab:adapter}. Note that all the approaches bring additional costs to get better generation results inevitably compared with Stable Diffusion. Thus our costs are acceptable when considering the significant improvement we contribute.

\noindent\textbf{Impact on Controllability with Different Methods.}~
In our experiments, both our method and A\&E managed to generate additional objects, but sometimes struggled with control precision, as seen with A\&E's canny controls in Figure~\ref{fig-exp-dominance}. We followed controllability evaluations from~\cite{zhao2023unicontrolnet} and assessed control effectiveness by calculating L2 distances from generated image conditions, and CLIP scores for object placement with GLIGEN. Despite occasional challenges, our method showed lower relative image-condition distances in Table~\ref{tab-image-condition}, indicating better control retention while adding objects.

\subsection{User Study}
To rigorously assess the effectiveness of our approach, we organized a comprehensive user study. Participants, numbering \textbf{81} in total, were shown 6 prompts randomly paired with various control conditions. This setup generated \textbf{10} unique scenarios where condition misalignment was a notable challenge. In each case, participants were presented with a set of 4 images synthesized by different methods under the same condition combinations. The core question of the survey was to identify which set of images most accurately reflected the intended combination of text and control image conditions. Remarkably, 71\% of the participants found that the images generated by our method best adhered to the specified conditions, as detailed in Table~\ref{tab-exp-user_study}. This overwhelming preference highlights our method's superior capability in effectively addressing and rectifying condition misalignments, thereby validating its practical applicability and user satisfaction in real-world scenarios.

\begin{table}[tb]
\caption{\textbf{User study results.}}
\centering
\resizebox{0.65\columnwidth}{!}{%
\begin{tabular}{ccccc}
\toprule
                        & SD~\cite{rombach2022high} & SSD~\cite{feng2022training} & AE~\cite{chefer2023attend}     & Ours  \\ \midrule
{User preference}     
                        & 11\%  &  7\% & 11\%  & \textbf{71\%} \\ \bottomrule
\end{tabular}%
}
\label{tab-exp-user_study}
\end{table}

\begin{table}[htb]
\caption{\textbf{Ablation Study.}}
\centering
\resizebox{0.8\columnwidth}{!}{%
\begin{tabular}{lcccc}
\toprule
            & \multicolumn{2}{c}{with ControlNets \cite{zhang2023adding}}             & \multicolumn{2}{c}{with GLIGEN \cite{li2023gligen}}   \\ \cmidrule(lr){2-3}\cmidrule(lr){4-5}
            & Simialrity$\uparrow$         &Distance$\downarrow$          & Simialrity$\uparrow$             & Distance$\downarrow$   \\ \midrule
        
SD          & 69.49\%            & -                & 71.24\%                & - \\
Ours-w/o-realign  & 73.78\%            & 9.59\%           & 74.81\%                & 6.23\% \\
Ours        & \textbf{73.88\%}   & \textbf{1.58\%}  & \textbf{79.87\%}       & \textbf{4.34\%} \\ \bottomrule
\end{tabular}%
}
\label{tab-ablation}
\end{table}

\subsection{Ablation Study}
\begin{figure}[!t]
    \centering
    \includegraphics[width=1.0\linewidth]{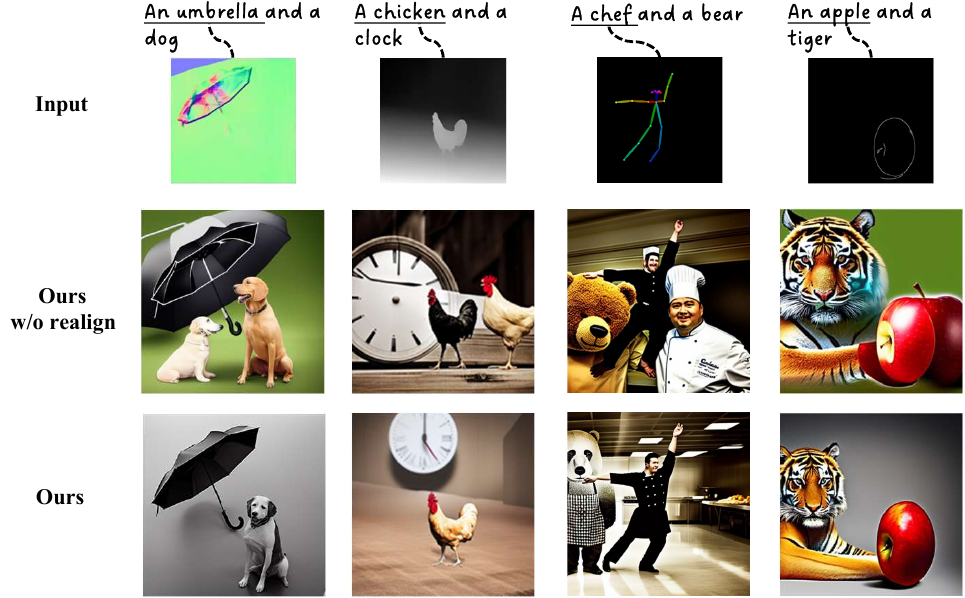}
    \caption{\textbf{Ablation Study.} Incorporating attention realignment operation into our method prevents the generation of extraneous objects, as exemplified by the absence of redundant elements like the second chef in the column of ``pose''.
    }
    \label{fig-exp-ablation}
\end{figure}

We carried out ablation studies using Stable Diffusion with ControlNets or GLIGEN. The results, presented in Table~\ref{tab-ablation} and Figure~\ref{fig-exp-ablation}, reveal that our TASC significantly boosts image-text similarity, effectively generating an additional object amidst dominance. Note that the ``ours-w/o-realign'' method can be considered an enhanced version of Composable Diffusion \cite{liu2022compositional}, effectively eliminating any condition misalignment issues in all scoring computations.
However, excluding attention realignment operation led to issues in object generation under image conditions, as shown in Figure~\ref{fig-exp-ablation}, and resulted in poorer performance in relative image-condition distance. Our full method excels in balancing relative image-condition distances and image-text similarity, producing images that adeptly meet both text and image condition constraints, demonstrating a harmonized approach.

\begin{figure}[!t]
    \centering
    \includegraphics[width=1.0\linewidth]{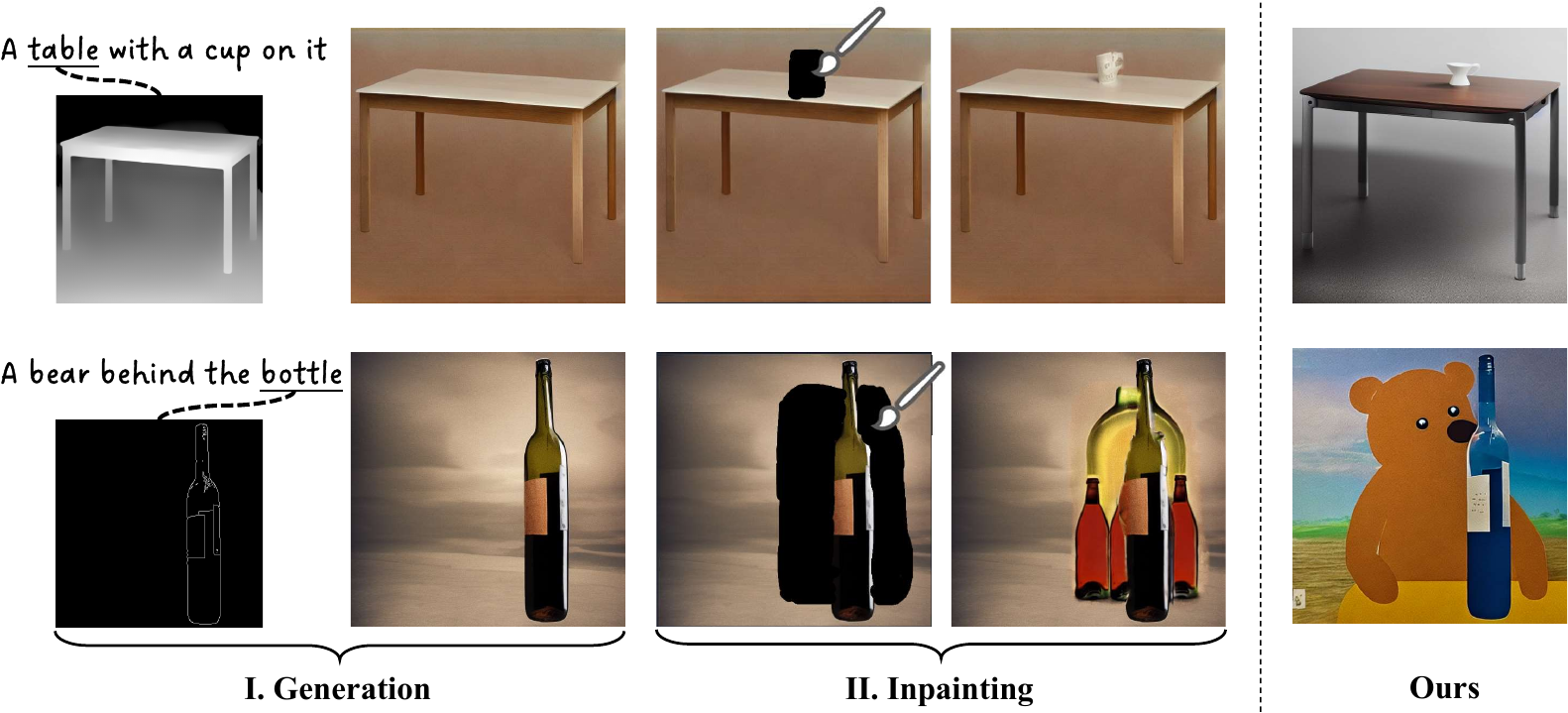}
    \caption{\textbf{Left: Two-stage method. Right: Our Approach.} Our method autonomously positions objects from the text in their correct locations, leveraging the generative model's inherent priors, without the need for manual intervention such as masking. Inpainting is performed using~\cite{avrahami2022blended}.}
    \label{fig-exp-model_prior}
\end{figure}

\subsection{Discussions}
Our method unifies multiple model consensuses through the cross-attention mechanism, ensuring that objects in the text are generated correctly according to model priors. This contrasts with stepwise generation approaches, as illustrated in Figure~\ref{fig-exp-model_prior}, which require manual intervention (like providing masks) and may struggle in complex scenarios (e.g., ``bear behind bottle''). Our approach effectively leverages model priors.
It is noted that the effectiveness of our "Realign" strategy is limited by the model's cross-attention control capability. If the model has a bias in understanding certain words (where the attention map does not fully correspond to the correct text), the alignment efficacy using words as bridges is impacted.

\section{Conclusion and Limitation}
\label{sec:conclusion}
In summary, our method significantly improves the control capabilities of text-to-image diffusion models. It adeptly manages unaligned conditions, outperforming recent methods in efficiency and effectiveness. This training-free, two-phase approach allows for more intricate and flexible image synthesis, advancing the domain of controllable image generation. As for limitations, our method brings additional computation costs like other state-of-the-art methods and may limit the practical usage to a certain degree.

\section*{Acknowledgments}
This work is supported by National Natural Science Foundation of China (No. 62206068) and Guangzhou-HKUST(GZ) Joint Funding Scheme (No. 2024A03J0241). Finally, we extend our gratitude to the anonymous reviewers and conference organizers for their constructive comments and efforts towards improving this manuscript.

%
%
\bibliographystyle{splncs04}
\bibliography{main}

\begin{thebibliography}{10}
\providecommand{\url}[1]{\texttt{#1}}
\providecommand{\urlprefix}{URL }
\providecommand{\doi}[1]{https://doi.org/#1}

\bibitem{avrahami2022blended}
Avrahami, O., Lischinski, D., Fried, O.: Blended diffusion for text-driven editing of natural images. In: Proceedings of the IEEE/CVF Conference on Computer Vision and Pattern Recognition. pp. 18208--18218 (2022)

\bibitem{bhunia2022person}
Bhunia, A.K., Khan, S., Cholakkal, H., Anwer, R.M., Laaksonen, J., Shah, M., Khan, F.S.: Person image synthesis via denoising diffusion model. arXiv preprint arXiv:2211.12500  (2022)

\bibitem{brack2023sega}
Brack, M., Friedrich, F., Hintersdorf, D., Struppek, L., Schramowski, P., Kersting, K.: Sega: Instructing diffusion using semantic dimensions. arXiv preprint arXiv:2301.12247  (2023)

\bibitem{brooks2022instructpix2pix}
Brooks, T., Holynski, A., Efros, A.A.: Instructpix2pix: Learning to follow image editing instructions. arXiv preprint arXiv:2211.09800  (2022)

\bibitem{chefer2023attend}
Chefer, H., Alaluf, Y., Vinker, Y., Wolf, L., Cohen-Or, D.: Attend-and-excite: Attention-based semantic guidance for text-to-image diffusion models. arXiv preprint arXiv:2301.13826  (2023)

\bibitem{chen2024training}
Chen, M., Laina, I., Vedaldi, A.: Training-free layout control with cross-attention guidance. In: Proceedings of the IEEE/CVF Winter Conference on Applications of Computer Vision. pp. 5343--5353 (2024)

\bibitem{choi2021ilvr}
Choi, J., Kim, S., Jeong, Y., Gwon, Y., Yoon, S.: Ilvr: Conditioning method for denoising diffusion probabilistic models. arXiv preprint arXiv:2108.02938  (2021)

\bibitem{dhariwal2021guided}
Dhariwal, P., Nichol, A.: Diffusion models beat gans on image synthesis  (2021)

\bibitem{feng2022training}
Feng, W., He, X., Fu, T.J., Jampani, V., Akula, A., Narayana, P., Basu, S., Wang, X.E., Wang, W.Y.: Training-free structured diffusion guidance for compositional text-to-image synthesis. arXiv preprint arXiv:2212.05032  (2022)

\bibitem{gal2022image}
Gal, R., Alaluf, Y., Atzmon, Y., Patashnik, O., Bermano, A.H., Chechik, G., Cohen-Or, D.: An image is worth one word: Personalizing text-to-image generation using textual inversion. arXiv preprint arXiv:2208.01618  (2022)

\bibitem{ge2023expressive}
Ge, S., Park, T., Zhu, J.Y., Huang, J.B.: Expressive text-to-image generation with rich text (2023)

\bibitem{hertz2022prompt}
Hertz, A., Mokady, R., Tenenbaum, J., Aberman, K., Pritch, Y., Cohen-Or, D.: Prompt-to-prompt image editing with cross attention control. arXiv preprint arXiv:2208.01626  (2022)

\bibitem{heusel2017gans}
Heusel, M., Ramsauer, H., Unterthiner, T., Nessler, B., Hochreiter, S.: Gans trained by a two time-scale update rule converge to a local nash equilibrium. Advances in neural information processing systems  \textbf{30} (2017)

\bibitem{ho2020denoising}
Ho, J., Jain, A., Abbeel, P.: Denoising diffusion probabilistic models. NIPS  (2020)

\bibitem{ho2022classifier}
Ho, J., Salimans, T.: Classifier-free diffusion guidance. arXiv preprint arXiv:2207.12598  (2022)

\bibitem{hu2021lora}
Hu, E.J., Shen, Y., Wallis, P., Allen-Zhu, Z., Li, Y., Wang, S., Wang, L., Chen, W.: Lora: Low-rank adaptation of large language models. arXiv preprint arXiv:2106.09685  (2021)

\bibitem{huang2023reversion}
Huang, Z., Wu, T., Jiang, Y., Chan, K.C., Liu, Z.: Reversion: Diffusion-based relation inversion from images. arXiv preprint arXiv:2303.13495  (2023)

\bibitem{kim2023dense}
Kim, Y., Lee, J., Kim, J.H., Ha, J.W., Zhu, J.Y.: Dense text-to-image generation with attention modulation. In: Proceedings of the IEEE/CVF International Conference on Computer Vision. pp. 7701--7711 (2023)

\bibitem{li2022blip}
Li, J., Li, D., Xiong, C., Hoi, S.: Blip: Bootstrapping language-image pre-training for unified vision-language understanding and generation. In: ICML (2022)

\bibitem{li2023gligen}
Li, Y., Liu, H., Wu, Q., Mu, F., Yang, J., Gao, J., Li, C., Lee, Y.J.: Gligen: Open-set grounded text-to-image generation. In: CVPR. pp. 22511--22521 (2023)

\bibitem{liang2023luciddreamer}
Liang, Y., Yang, X., Lin, J., Li, H., Xu, X., Chen, Y.: Luciddreamer: Towards high-fidelity text-to-3d generation via interval score matching (2023)

\bibitem{lin2014microsoft}
Lin, T.Y., Maire, M., Belongie, S., Hays, J., Perona, P., Ramanan, D., Doll{\'a}r, P., Zitnick, C.L.: Microsoft coco: Common objects in context. In: Computer Vision--ECCV 2014: 13th European Conference, Zurich, Switzerland, September 6-12, 2014, Proceedings, Part V 13. pp. 740--755. Springer (2014)

\bibitem{liu2022compositional}
Liu, N., Li, S., Du, Y., Torralba, A., Tenenbaum, J.B.: Compositional visual generation with composable diffusion models. In: Computer Vision--ECCV 2022: 17th European Conference, Tel Aviv, Israel, October 23--27, 2022, Proceedings, Part XVII. pp. 423--439. Springer (2022)

\bibitem{meng2021sdedit}
Meng, C., He, Y., Song, Y., Song, J., Wu, J., Zhu, J.Y., Ermon, S.: Sdedit: Guided image synthesis and editing with stochastic differential equations. In: International Conference on Learning Representations (2021)

\bibitem{mokady2022null}
Mokady, R., Hertz, A., Aberman, K., Pritch, Y., Cohen-Or, D.: Null-text inversion for editing real images using guided diffusion models. arXiv preprint arXiv:2211.09794  (2022)

\bibitem{mou2023t2i}
Mou, C., Wang, X., Xie, L., Zhang, J., Qi, Z., Shan, Y., Qie, X.: T2i-adapter: Learning adapters to dig out more controllable ability for text-to-image diffusion models. arXiv preprint arXiv:2302.08453  (2023)

\bibitem{nichol2021glide}
Nichol, A., Dhariwal, P., Ramesh, A., Shyam, P., Mishkin, P., McGrew, B., Sutskever, I., Chen, M.: Glide: Towards photorealistic image generation and editing with text-guided diffusion models. arXiv preprint arXiv:2112.10741  (2021)

\bibitem{parmar2023zero}
Parmar, G., Singh, K.K., Zhang, R., Li, Y., Lu, J., Zhu, J.Y.: Zero-shot image-to-image translation. arXiv preprint arXiv:2302.03027  (2023)

\bibitem{poole2022dreamfusion}
Poole, B., Jain, A., Barron, J.T., Mildenhall, B.: Dreamfusion: Text-to-3d using 2d diffusion. arXiv preprint arXiv:2209.14988  (2022)

\bibitem{ramesh2022dalle2}
Ramesh, A., Dhariwal, P., Nichol, A., Chu, C., Chen, M.: Hierarchical text-conditional image generation with clip latents. arXiv preprint arXiv:2204.06125  (2022)

\bibitem{rombach2022high}
Rombach, R., Blattmann, A., Lorenz, D., Esser, P., Ommer, B.: High-resolution image synthesis with latent diffusion models. In: CVPR (2022)

\bibitem{ruiz2022dreambooth}
Ruiz, N., Li, Y., Jampani, V., Pritch, Y., Rubinstein, M., Aberman, K.: Dreambooth: Fine tuning text-to-image diffusion models for subject-driven generation. arXiv preprint arXiv:2208.12242  (2022)

\bibitem{saharia2022imagen}
Saharia, C., Chan, W., Saxena, S., Li, L., Whang, J., Denton, E., Ghasemipour, S.K.S., Ayan, B.K., Mahdavi, S.S., Lopes, R.G., et~al.: Photorealistic text-to-image diffusion models with deep language understanding. arXiv preprint arXiv:2205.11487  (2022)

\bibitem{sohl2015deep}
Sohl-Dickstein, J., Weiss, E., Maheswaranathan, N., Ganguli, S.: Deep unsupervised learning using nonequilibrium thermodynamics. In: International Conference on Machine Learning. PMLR (2015)

\bibitem{song2020ddim}
Song, J., Meng, C., Ermon, S.: Denoising diffusion implicit models. arXiv preprint arXiv:2010.02502  (2020)

\bibitem{song2019score1}
Song, Y., Ermon, S.: Generative modeling by estimating gradients of the data distribution. Advances in Neural Information Processing Systems  \textbf{32} (2019)

\bibitem{song2020score2}
Song, Y., Sohl-Dickstein, J., Kingma, D.P., Kumar, A., Ermon, S., Poole, B.: Score-based generative modeling through stochastic differential equations. arXiv preprint arXiv:2011.13456  (2020)

\bibitem{tumanyan2022plug}
Tumanyan, N., Geyer, M., Bagon, S., Dekel, T.: Plug-and-play diffusion features for text-driven image-to-image translation. arXiv preprint arXiv:2211.12572  (2022)

\bibitem{wang2023modelscope}
Wang, J., Yuan, H., Chen, D., Zhang, Y., Wang, X., Zhang, S.: Modelscope text-to-video technical report. arXiv preprint arXiv:2308.06571  (2023)

\bibitem{wang2023not}
Wang, L., Yang, S., Liu, S., Chen, Y.c.: Not all steps are created equal: Selective diffusion distillation for image manipulation. In: Proceedings of the IEEE/CVF International Conference on Computer Vision. pp. 7472--7481 (2023)

\bibitem{wang2024instancediffusion}
Wang, X., Darrell, T., Rambhatla, S.S., Girdhar, R., Misra, I.: Instancediffusion: Instance-level control for image generation. In: Proceedings of the IEEE/CVF Conference on Computer Vision and Pattern Recognition. pp. 6232--6242 (2024)

\bibitem{xie2023boxdiff}
Xie, J., Li, Y., Huang, Y., Liu, H., Zhang, W., Zheng, Y., Shou, M.Z.: Boxdiff: Text-to-image synthesis with training-free box-constrained diffusion. In: Proceedings of the IEEE/CVF International Conference on Computer Vision. pp. 7452--7461 (2023)

\bibitem{yang2023denoising}
Yang, S., Chen, Y., Wang, L., Liu, S., Chen, Y.: Denoising diffusion step-aware models. arXiv preprint arXiv:2310.03337  (2023)

\bibitem{zeng2022scenecomposer}
Zeng, Y., Lin, Z., Zhang, J., Liu, Q., Collomosse, J., Kuen, J., Patel, V.M.: Scenecomposer: Any-level semantic image synthesis. arXiv preprint arXiv:2211.11742  (2022)

\bibitem{zhang2023adding}
Zhang, L., Agrawala, M.: Adding conditional control to text-to-image diffusion models. arXiv preprint arXiv:2302.05543  (2023)

\bibitem{zhao2022egsde}
Zhao, M., Bao, F., Li, C., Zhu, J.: Egsde: Unpaired image-to-image translation via energy-guided stochastic differential equations. arXiv preprint arXiv:2207.06635  (2022)

\bibitem{zhao2023unicontrolnet}
Zhao, S., Chen, D., Chen, Y.C., Bao, J., Hao, S., Yuan, L., Wong, K.Y.K.: Uni-controlnet: All-in-one control to text-to-image diffusion models (2023)

\end{thebibliography}
\end{document}